# Model Robustness with Text Classification: Semantic-preserving adversarial attacks

Rahul Singh, Tarun Joshi, Vijayan N. Nair, and Agus Sudjianto
Corporate Model Risk, Wells Fargo, USA

## Abstract

We propose algorithms to create adversarial attacks to assess model robustness in text classification problems. They can be used to create white box attacks and black box attacks while at the same time preserving the semantics and syntax of the original text. The attacks cause significant number of flips in white-box setting and same rule based can be used in black-box setting. In a black-box setting, the attacks created are able to reverse decisions of transformer based architectures.

## 1 Introduction and Literature Review

Training data, on which machine learning models are developed, are sometimes not representative of the datasets of interest or future datasets. This causes the models to yield poor predictions on the unseen data. It has been seen in the literature [1] that minute perturbations in the training data can cause the model's performance to drop significantly. By systematically creating such perturbations, one can assess and create robust models through a process called adversarial training [1]. Adversarial training can help us learn the behavior of our models and the decision boundaries where breakdown in robustness occurs.

Similar concerns related to model robustness have also been investigated in Natural Language Processing (NLP). Some of the papers that study these perturbations include characters level changes [2], word level changes [3, 4], and sentence level changes [5]. Table 1 shows selected examples of adversarial attacks (perturbations) and the corresponding degradation in model performance (classification as negative). For example, replacing the word "good" by "fantastic" changes the probability of being classified as negative from 0.56 to 0.33. Similarly, adding the word "highly" in front of "trustworthy" causes a drastic decrease in the probability from 0.91 to 0.48.

Along with this, different NLP applications have also been tested, including Text Classification [2, 4], Machine Translation [5], and Question Answering. [6] The collection of research related to adversarial learning can be broadly divided into two categories, with respect to how adversaries are created and the amount of model information available. If adversarial inputs are created by using the model information, they are called "white box attacks". Perhaps the more interesting category is one in which nothing is known about the model. These are called as "black box attacks". We will also refer to them as "supervised and unsupervised perturbations" of the data.

Research in adversarial attacks for deep learning models started with the introduction of a brute force attack to change the model prediction [7]. Since then, research in this field has exploded and many different types of attacks have been explored. Adversaries are created using different definitions of distance between the original input and perturbed input, including $L_0$, $L_1$ and $L_2$ norms [8]. While there are many types of perturbations including random ones, the most common white-box attacks are gradient-based [1] that use projected gradient information in the model to determine the "best" perturbation and create adversaries.

Table 1: Two examples of model non-robustness behavior observed by creating adversarial attacks from the rule-based trigger algorithm. The words highlighted with color yellow and green represent the original word and its replacement respectively. The probability values of the model (represent the model's probability p(y|x)) for both the original and perturbed messages

| Original Text | Adversarial Text |
| --- | --- |
| The device is easy to use, but selection of a station to listen to with **good** reception is difficult. | The device is easy to use, but selection of a station to listen to with **fantastic** reception is difficult. |
| Probability = **0.556183** | Probability = **0.325465** |
| Disc after disc, burn after burn, Sony CD-R's just. don't. work. If nothing else, they're consistent. Why doesn't anybody sell **trustworthy** products anymore? | Disc after disc, burn after burn, Sony CD-R's just. don't. work. If nothing else, they're consistent. Why doesn't anybody sell **highly trustworthy** products anymore? |
| Probability = **0.912351** | Probability = **0.475465** |

These and similar ideas in the field of adversarial training have also been used in the NLP community for assessing robustness of models for unstructured data. These include strategies that apply insertion, modification, and removal of characters [2]. The actual choices of words or characters to perturb are selected using the highest gradient of the cost (loss) function. These can be directed attacks in which changes are only focused on certain types of words based on their Parts-of-Speech (PoS) tag. Examples include: i) only adverbs with the highest gradient for creating perturbations and insertion of text in the document [3]; ii) Jacobian based saliency attack [9] to find words that have the same sign in the embedding space as the sign of the Jacobian of the outputs from the final layer of the model before the final decision; iii) character level changes that cause maximum perturbations using derivatives as a surrogate loss for character based models and this method can be extended to word level changes [4].

There have also been development of black box attacks. Semantically equivalent adversarial rules (SeAR) were successfully created by using only paraphrase modeling and the final model decision information [10]. Genetic algorithms have been applied along with paraphrase models to create diverse adversaries. [11] One recent algorithm even explores the concept of universal triggers that, when added to every text instance, can alter the model behavior significantly. [12]

This paper presents novel algorithms to perform adversarial testing of NLP text classification models. The results can be easily extended to other NLP tasks. The main contributions of the paper are:

- Methods to create both white-box and black-box attacks for NLP text classification problems.

- Two new algorithms that can be used with any deep learning model to create a wide range of adversaries.
- A sequence of checks (or steps) to maintain the semantic nature of the adversaries relative to the original text. We use a combination of embedding information, semantic polarity, PoS tagging information and mask language model predictions to keep the semantic information intact.

We also show that the adversaries that are created can be used to develop general rules about the data, which can later be used for black box models.

Table 2 provides a collection of abbreviations and their definitions. These will be used throughout the paper.

Table 2: Abbreviations that will be used frequently in the paper and their definitions

| Abbreviations | Definitions |
|---|---|
| Adj: | Adjective |
| NN: | Noun |
| Adv: | Adverb |
| VB: | Verb |
| $t_{emb}$: | Minimum value of cosine similarity of embedding vectors of two words |
| $t_{bert}$: | Minimum value of prediction probability of missing words using BERT [13] |
| POS: | Parts of Speech |

The paper is organized as follows. Section 2 describes the algorithms and different strategies related to both white-box and black-box settings. In Section 3, we discuss the experiments performed showing the effect of algorithm on different machine learning models. We extend this with some discussion on the experimental results and in Section 4, we conclude the paper with some insights and direction for future work.

# 2 Attacks and algorithms

We introduce a rule-based algorithm to create white box and black box attacks that do not change the semantics of the document. The white box attacks are categorized into two strategies as Replacement and Insertion. In the Replacement strategy, we replace some tokens in the targeted dataset, while for Insertion, we insert some tokens in the dataset. The black box attacks are designed by using common patterns from the replacement strategy of the white box attacks. We collect some common rules from the replacement strategy and apply those rules to create black box attacks.

The first step in both replacement and insertion strategies is to find universal triggers [12] constrained by predefined rules. We apply universal trigger algorithm to find a list of triggers that increase overall loss for the test sample (i.e. find triggers which when applied in isolation causes the loss to increase and thus indicate a potential towards model flipping behavior). Specifically, we use a projected gradient descent method to find the list of universal triggers (or tokens) in the embedding space that affect the model decision the most. Specifically, consider a word $w^j$ and let $e^j$ denote its word embedding. Let $L$ be a suitable loss function and $\nabla_{e^j} L$ be its projected gradient. We use an iterative method in the positive gradient direction to find the appropriate word trigger, subject to the perturbation being constrained to be in embedding space. Given a vector $e_i^j$, we find the next word ($e_{i+1}^j$) using the dot product of ($e_{i+1}^j - e_i^j$) with the gradient ($\nabla_{e_i^j} L$) vector. We take steps in the direction of gradient (of loss) with the additional condition that they must satisfy some predefined rules related to the sequence of POS tag of the tokens. This algorithm, modified form of universal trigger algorithm [12], is called the global strategy and is described below:

I. Initialize the trigger sequence by using the words like "the", the character "a", or the sub-word "an" and concatenate the trigger to the front or end of all inputs.
II. Replace the trigger sequence using rule based search over the embedding space and find the tokens by the following equation:

$$\underset{e_{i+1}^j \in V}{\mathrm{argmin}} \left[ e_{i+1}^j - e_i^j \right] \nabla_{e_i^j} L$$

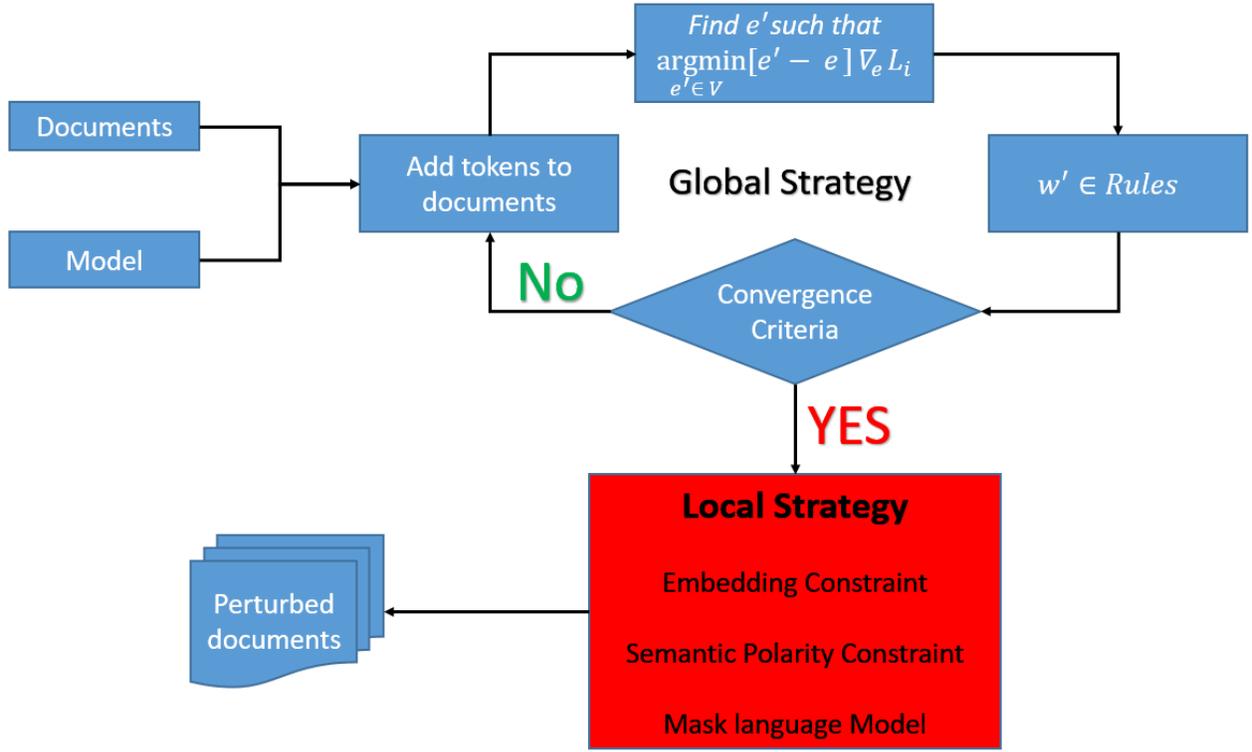

Figure 1: Schematic diagram of the algorithms introduced in this paper. $L_i$, $e'$, $w'$, $V$ represent loss, embedding vector for a word, selected word, and vocabulary of the dataset respectively

$subject\ to\ the\ constratint\ POS_{tag}(trigger\ sequence(w^1 \ldots w^j..)) = Rules$

Here, we take $L$ to be the logistic loss function: $L = -y\log\hat{y} - (1-y)\log(1-\hat{y})$;
$i$ is the step in iteration;

$w^j$ is the *j*-th word in the trigger sequence;
$e_i^j$ represents the embedding vector of word, $w^j$;
*y* is the original label; and
*ŷ* is the model prediction

III. Find the word in the embedding space that satisfies the above conditions.
IV. Apply rule-based search to find the best collection of tags that maximize the loss.
V. Repeat steps (II - IV) until we reach maximum iterations or there are no more words in the vocabulary space that satisfy the conditions specified in II.

## 2.1 Replacement – Create semantically equivalent triggers by replacing tokens

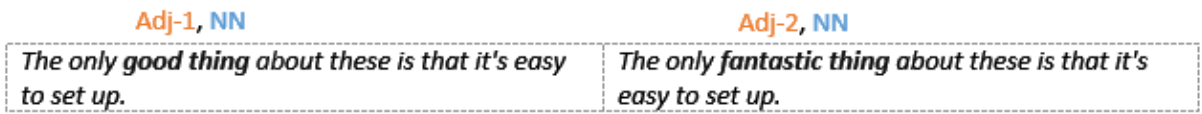

Figure 2: An example of perturbation using Replacement strategy. An adjective (good) in a (Adjective, Noun) pair is replaced with one of the trigger adjectives (fantastic)

In the *replacement* strategy, we find triggers that will replace adjectives in the text, i.e., (POS(trigger) = adjective), using the global strategy described in the algorithm above. We find semantically equivalent triggers that change the model decision. The steps are as follows:

1. Perform PoS tagging for each message in the target dataset.
2. Find all (*adj, NN*) pairs in the message (PoS tagging). For example in the sentence below:
   Example Message:
   *Sound stinks. If it weren't for the hassle of returning items I would return these. The only **good thing** about these is that it's easy to set up. You can't listen to any of your other speakers while you have these plugged in. They go into the headphone outlet of the stereo. I bought these to have **additional sound** outside. If you have an MP3 player and want to use it for that, it will probably work out better for you. But again the sound isn't all that great.*
   (adj, NN) pairs – **(good thing), (additional sound)**
3. **[Candidate Selection]** Derive a list of potential triggers for the adjective part using the universal triggers. We apply the following rules:
   a. The adjective in the message and adjective in the trigger should have same polarity.
   b. The cosine similarity between the two should be greater than a pre-defined threshold ($t_{emb}$) in the embedding space. In our experiments, we used a threshold of 0.45 ($t_{emb}$ = 0.45) and GloVe embeddings [14] to represent words in the embedding space.
   c. Sort the list of potential triggers in order of decreasing cosine similarity.
   d. Example
      i. ('additional', 'sound'), List of options -> ['several', 'numerous', 'full', 'possible', 'extensive', 'many', 'related', 'total', 'extra', 'few', 'able', 'potential', 'various', 'likely', 'individual', 'specific']
      ii. ('good', 'thing'), List of options -> ['strong', 'effective', 'successful', 'easy', 'impressed', 'clear', 'pleased', 'happy', 'important', 'comfortable', 'certain',

'useful', 'positive', 'fantastic', 'solid', 'healthy', 'free', 'nice', 'great', 'safe', 'true']
   e. **[Candidate Replacement]** Replace an adjective with each trigger and test if the message flips. For example, an (adj, NN) trigger is (additional, sound) and [many, extra, full, total] is a list of potential triggers which can replace "additional" in the original message to flip the model prediction.
   f. BERT masked Language model is used to find the possible candidates which retain syntax and semantics of the sentence. We use the BERT prediction of missing words and select words whose prediction probability is greater than $t_{BERT}$. We use a $t_{BERT}$ equal to 1e-3 in this experiment. An example is shown in Table 3.

Table 3: Words with BERT prediction probabilities greater than $t_{BERT}$ are selected

| Input sentence | Input options | BERT prediction probability | Selected |
|---|---|---|---|
| I bought these to have **[MASK]** sound outside | many | 3.76e-05 | No |
| | extra | 0.01442 | YES |
| | full | 0.00021 | No |
| | total | 5.21e-05 | No |

We have also tested other rules for replacement including (Adverb, Adjective), (Noun, Verb). This is a general strategy in which other rules can also be included easily.

## 2.2 Insertion – Create semantically equivalent triggers by inserting tokens

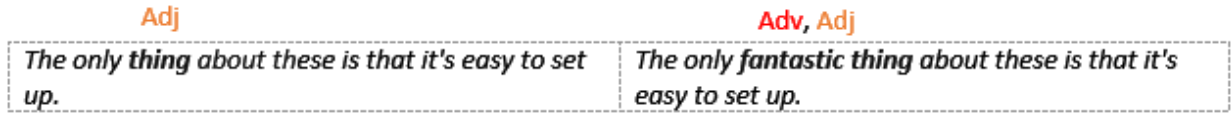

Figure 3: An example of perturbation using Insertion strategy. It shows an adverb (fantastic) is inserted before an adjective (thing) using the Insertion strategy

For insertion, we use the global strategy to find triggers that follow the Rule = POS(trigger1, trigger2) = (Adverb, Adjective). In the description below, we illustrate only the Adjective part of the trigger to find the semantically equivalent adjectives in the dataset. The steps are similar to the Replacement strategy with a few minor differences:

1. Find all *adj* in the message (PoS tagging).
2. **[Candidate Selection]** Derive a list of potential triggers for the adjective part using the universal triggers by applying the following rules:
   a. The adjective in message and adjective in trigger should have the same polarity.
   b. The cosine similarity between the two should be greater than the embedding threshold ($t_{emb}$). In our experiments, we used ($t_{emb}$ = 0.45) and GloVe embeddings to represent words in the embedding space.
   c. Sort the list of potential triggers in order of decreasing cosine similarity.

d. **[Candidate Insertion]** Insert the adverb from the trigger part of adverb in front of the candidate adjectives and test if the message flips.
e. BERT masked Language model is used to filter the candidates which retain syntax and semantics of the sentence.

For insertion, other strategies following different rules can also be used with the same steps described above as long as we maintain the semantics of the sentence.

## 2.3 Black-box attack – Rules to attack black box models

Table 4: Black-box rules obtained from the Replacement strategy that successfully flipped the model decision

| Black box rules (Adj, NN) | | |
|---|---|---|
| Original Pair | | Replaced Pair |
| good [Noun] | → | fantastic [Noun] |
| nice [Noun] | → | good [Noun] |
| first [Noun] | → | second [Noun] |
| small [Noun] | → | tiny [Noun] |
| smooth [Noun] | → | soft [Noun] |

In this section, we collect all the rules that were successful in changing the decision of the trained model. We generalize these rules and apply to the same dataset and test them on other text classification models. A rule is comprised of pair of words and takes the form, for example, r = [(Noun, b) → (Noun, c)], where b is replaced by c for every instance that includes (Noun, b) as shown in Table 4. The output of every text after applying rule r changes all instances of (Noun, b) to (Noun, c). Table 4 presents some of the rules that are used and applied to the sentences. The rule, r = [(good, Noun) → (fantastic, Noun)] changes all (good, Noun) pairs in the sentences to (fantastic, Noun) and similarly other rules are applied to all the sentences.

Table 5 presents one example of semantic-preserving adversaries by applying the three strategies, Replacement, Insertion and the rule [good, Noun) → (fantastic, Noun)]. In all the shown cases, the model decision changes from negative review to positive review although the complete meaning of the sentence is not changed. The "New" sentences still maintain the negative polarity that represent the negative review and would be difficult for a human to differentiate from the original message.

# 3 Experiments
## 3.1.1 Dataset
The two strategies in Sections 3.1 and 3.2 can be applied to any differentiable deep learning (DL) model. Here we focus on three different DL models for classifying text documents. The first is a Convolution Neural Network (CNN) introduced in [15] for text classification, the second model is a stacked two-layer LSTM model, and the last one is DistillBERT, a transformer architecture [16].These are popular machine learning algorithms used for text classification problems and to extract features from the dataset in different ways. We apply the first two strategies on CNN networks and use the results to derive general rules. These rules are then used as a black-box attack for the other two models (i.e. 2-layer LSTM and DistillBERT). More details related to the CNN model are provided in Table 6. The LSTM model is a two-layer bidirectional model with 100 hidden layer units.

Table 5: Examples of a reversed model decision by using the three strategies, Replacement, Insertion and black-box attack using the rule (good [NN] -> fantastic [NN])

| Methods | Previous Message | New Message |
|---|---|---|
| Replacement | Sound stinks. If it weren't for the hastle of returning items I would return these. The only **good** thing about these is that it's easy to set up. You can't listen to any of your other speakers while you have these plugged in. They go into the headphone outlet of the stereo. I bought these to have additional sound outside. If you have an MP3 player and want to use it for that, it will probably work out better for you. But again the sound isn't all that great.<br><br>Probability = **0.631151** | Sound stinks. If it weren't for the hastle of returning items I would return these. The only **useful** thing about these is that it's easy to set up. You can't listen to any of your other speakers while you have these plugged in. They go into the headphone outlet of the stereo. I bought these to have additional sound outside. If you have an MP3 player and want to use it for that, it will probably work out better for you. But again the sound isn't all that great.<br><br>Probability = **0.472134** |
| Insertion | Tele Type WorldNav 3100 Deluxe GPS is not **beneficial** GPS for the navigation because it doesnot provide the basic services such as Transportation like Taxis, Amtrak, Recreation, etc. So I can rate this product 0000000000000.<br><br>Probability = **0.758428** | Tele Type WorldNav 3100 Deluxe GPS is not **enormously beneficial** GPS for the navigation because it doesnot provide the basic services such as Transportation like Taxis, Amtrak, Recreation, etc. So I can rate this product 0000000000000.<br><br>Probability = **0.351217** |
| Semantic rules<br><br>good [Noun] => fantastic [Noun] | The device is easy to use, but selection of a station to listen to with **good** reception is difficult. when driving in rural areas, can get some reception for a while. but need to change stations frequently to maintain reception. in city areas, very poor reception. susceptible to much interference. i like the design, but there is too much static on the reception to make listening enjoyable.<br><br>Probability = **0.556183** | The device is easy to use, but selection of a station to listen to with **fantastic** reception is difficult. when driving in rural areas, can get some reception for a while. but need to change stations frequently to maintain reception. in city areas, very poor reception. susceptible to much interference. i like the design, but there is too much static on the reception to make listening enjoyable.<br><br>Probability = **0.325465** |

CNN and LSTM models were trained with Adam optimizer [17] along with gradient clipping. DistillBERT has half the number of layers compared to the 12 layers of BERT-small. [13] The remaining implementation details are same as BERT. The contextual nature of transformer architecture is assumed to create a more robust model as it has less dependence on independent contributions. We use the dataset on electronics product reviews from Amazon [18] which is a subset of a large Amazon review dataset (see http://riejohnson.com/ cnn_data.html). See Table 7 for more details.

Following an earlier setup, [19], we use only the text section and ignore the summary section. We also consider only positive and negative reviews. More details are provided in Table 7. The machine learning models and adversarial perturbations discussed in the paper were developed using PyTorch [20] framework and huggingface transformer library [21].

### 3.1.2 Sentiment Analysis

This section illustrates the utility of rule-based semantically equivalent triggers with application to sentiment analysis. We show both white box attack and black box attack for the Amazon review dataset

(Table 7). In the first case we apply the algorithm on a CNN model and collect the rules as described above. The rules are then applied to trained LSTM and DistillBERT models. For comparison, note that the original accuracies of the trained CNN, LSTM and DistillBERT models on the test dataset are 90.1, 89.0 and 93.1 respectively.

Table 6: Parameters of machine learning models

| General Model Parameters | |
|---|---|
| Vocabulary size | 68,218 |
| Embedding dimension | 300 |
| Dropout rate | 0.2 |
| Maximum Sequence length | 256 |
| **CNN Model** | |
| Hidden layers | 1(10 neurons) |
| Filters | 1, 2, 3, 4 |
| Number of filters | 50 per filter |
| **LSTM Model** | |
| Hidden Dimension | 100 |
| LSTM layers | 2 |

Table 7: Information on Dataset

| Classes | Data Samples (per class) | Training data | Testing data |
|---|---|---|---|
| Two | • Positive 37472<br>• Negative 37472 | • Positive 18376<br>• Negative 18376 | • Positive 18376<br>• Negative 18376 |

### 3.1.3 White box Attacks

Table 8: Replacement results from different rule choices. The different rules, on average, cause a drop of ~3% in the accuracy of model

| Replacement – Changing words with similar words | | | | | | | |
|---|---|---|---|---|---|---|---|
| | **Original Accuracy** | $(Adv, Adj_1)$ → $(Adv, Adj_2)$ | $(Adv_1, Adj)$ → $(Adv_2, Adj)$ | $(Adj_1, NN)$ → $(Adj_2, NN)$ | $(Adj, NN_1)$ → $(Adj, NN_2)$ | $(VB_1, NN)$ → $(VB_2, NN)$ | $(VB, NN_1)$ → $(VB, NN_2)$ |
| Amazon | **90.1** | 87.6 | 87.9 | 86.2 | 87.2 | 88.7 | 89.2 |

Table 8 shows the results for six different replacement strategies to change words in the target dataset. For example, the rule (Adv, Adj1) → (Adv, Adj2) refers to the replacement of adjective in the pair. Similarly (Adv, $NN_1$) →(Adv, $NN_2$) refers to the replacement of nouns in the pair. We followed the procedure described above to maintain the semantic equivalence between the original sentence and the perturbed sentence (i.e. the sentence obtained after replacement in original sentence). See, for example, Table 5. The different replacement strategies result in degradation of the order of ~1-3%. -- changing from 90.1 to values ranging from 86.2 to 89.2. The ($Adj_1$, NN) → ($Adj_2$, NN) rule creates the maximum decrease from 90.1 to 86.2.

Table 9: Insertion strategy causes ~3% drop in accuracy. The two cases show an insertion of an adverb in front of an adjective and an insertion of an adjective after an adverb

| Insertion – Inserting words into the sentence | | | |
|---|---|---|---|
| | Original Accuracy | (Adj) → (Adv, Adj) | (Adv) →(Adv, Adj) |
| Amazon | **90.1** | 87.2 | 86.2 |

Table 9 shows results for the insertion strategy: inserting an adverb in front of adjective or inserting an adjective behind adverb. Both types of insertions cause an appreciable drop in the accuracy of the model of about ~3%.

### 3.1.4 Black box Attacks

For this case, we collected all the rules from the replacement strategy (Table 4) and applied them to two different models: LSTM model and DistillBERT. We also analyzed the performance changes by comparing the results with the amount of perturbation in the dataset. We quantified the perturbations with the number of words changed in the dataset: for example a single perturbation refers to a single word change locally. Table 10 shows the results. We can see that replacing one word does not affect DistillBERT results but decreases the performance of LSTM-based model by as much as ~4%. However, as we increase the number of perturbations, the performance of DistillBERT also starts to degrade and after four changes, the number of flips obtained from DistillBERT is similar to the flips obtained from white box attacks. The stable behavior of DistillBERT with respect to a single-word change can be attributed to the contextual nature of the transformer model. Single-word replacements does not cause a significant model flipping behavior because of the multi-head attention mechanism in transformer-based architectures as a single word representation is a contribution of surrounding words [22]. However, as more words are replaced, the performance of the model drops which is a concern for transformer based models.

Table 10: Comparison of black-box attack results on two different models, LSTM and DistillBERT with white-box attack result on CNN model

| Amazon dataset | | | |
|---|---|---|---|
| Number of changes | CNN | LSTM | DistillBERT |
| 0 | 90.1% | 89.1% | 93% |
| 1 | 87.4 (-2.7%) | 86.1 (-3%) | 92(-1%) |
| 2 | 85.9 (-4.2%) | 85.2(-4.1%) | 90.4(-2.6%) |
| 3 | 85.7 (-4.4%) | 84.9(-4.3%) | 88.4(-4.6%) |

# 4 Conclusions and Future Directions

We have presented three novel strategies to create semantically similar adversarial perturbations in the context of NLP problems. Although perturbations maintain the context and meaning of the text, they can cause, sometime significant, degradation in model performance.

Our approach can be used to create both white-box and black-box attacks, and it can be applied to any machine learning model. The rule-based approach presented also shows a sequence of steps to maintain the quality of perturbations, which can be easily be adapted to any perturbation scheme and used to maintain the semantics of the text.

Our experiments are limited, and more work is needed to quantify the full extent of model robustness for the different DL algorithms and types of attacks. In future work, we plan to apply this approach to a variety of text classification datasets and NLP tasks.